\documentclass[sigconf]{acmart}
\AtBeginDocument{%
  }

\setcopyright{acmlicensed}
\copyrightyear{2026}
\acmYear{2026}
\acmDOI{XXXXXXX.XXXXXXX}
\acmConference[EMF@HRI'26]{Workshop on Errors, Mistakes, and Failures in Humans and Robots at 2026 ACM/IEEE International Conference on Human-Robot Interaction}{March 16--19,
  2026}{Edinburgh, Scotland, UK}
\acmISBN{978-1-4503-XXXX-X/2018/06}

\begin{document}

\title{Achieving Interaction Fluidity in a Wizard-of-Oz Robotic System: A Prototype for Fluid Error-Correction}
\author{Carlos Baptista De Lima}
\email{c.v.baptistadelima@swansea.ac.uk}
\orcid{0000-0003-1518-364X}
\affiliation{
\institution{Swansea University}
\streetaddress{Bay Campus, Fabian Way}
\city{Swansea}
\country{United Kingdom}
\postcode{SA1 8EN}
}

\author{Julian Hough}
\email{julian.hough@swansea.ac.uk}
\orcid{0000-0002-4345-6759}
\affiliation{
\institution{Swansea University}
\streetaddress{Bay Campus, Fabian Way}
\city{Swansea}
\country{United Kingdom}
\postcode{SA1 8EN}
}

\author{Frank Förster}
\email{f.foerster@herts.ac.uk}
\orcid{0000-0003-1797-682X}
\affiliation{%
  \institution{University of Hertfordshire}
  \streetaddress{College Lane}
  \city{Hatfield}
  \country{United Kingdom}
  \postcode{AL10 9AB}
}

\author{Patrick Holthaus}
\email{p.holthaus@herts.ac.uk}
\orcid{0000-0001-8450-9362}
\affiliation{%
  \institution{University of Hertfordshire}
  \streetaddress{College Lane}
  \city{Hatfield}
  \country{United Kingdom}
  \postcode{AL10 9AB}
}

\author{Yongjun Zheng}
\email{y.zheng20@herts.ac.uk}
\orcid{0000-0003-0824-2365}
\affiliation{%
  \institution{University of Hertfordshire}
  \streetaddress{College Lane}
  \city{Hatfield}
  \country{United Kingdom}
  \postcode{AL10 9AB}
}

\begin{abstract}
Achieving truly fluid interaction with robots with speech interfaces remains a hard problem, and the experience of current Human-Robot Interaction (HRI) remains laboured and frustrating. Some of the barriers to fluid interaction stem from a lack of a suitable development platform for HRI for improving interaction, even in robotic Wizard-of-Oz (WoZ) modes of operation used for data collection and prototyping. Based on previous systems, we propose the properties of interruptibility and correction (IaC), pollability, latency measurement and optimisation and time-accurate reproducibility of actions from logging data as key criteria for a fluid WoZ system to support fluid error correction. We finish by presenting a Virtual Reality (VR) HRI simulation environment for mobile manipulators which meets these criteria.
\end{abstract}

\ccsdesc[500]{Human-centered computing~HCI theory, concepts and models}
\ccsdesc[300]{Computing methodologies~Robotic planning}
\ccsdesc[500]{Computer systems organization~External interfaces for robotics}

\keywords{Human-Robot Interaction, Fluidity, Virtual Reality Environments, Wizard-of-Oz}

\maketitle

\section{Introduction}

The challenge of achieving fluid interaction remains prevalent within the field of human-robot interaction (HRI). Despite advances in robot vision, motion, manipulation and automatic speech recognition, state-of-the-art HRI is frequently slow, laboured, and fragile. Improving the feeling of fluidity is a difficult task to achieve, with a variety of challenges from the robot control algorithms to the physical constraints of the robots \cite{hough2016investigating,Holthaus2023,abramsetal25iros}.

While fluidity has not been formally defined, it is felt by humans \cite{hough2016investigating}. We define fluidity and fluid interaction as the ability to seamlessly transition in turn-taking to allow appropriate overlap of turns between agents, including multimodally, and to allow action using prediction. For example, in human-human pick-and-place situations, an instruction like ``put the remote on the table" with a repair like ``no, the left-hand table" typically has no delay in reacting to the initial instruction, and adaptation to the correction is instant. Fluid interaction for robots consists of more seamless, human-like transitions from processing speech to taking physical actions with no delays, permitting appropriate overlap between the two, and the ability to repair actions in real time as humans do \cite{baptista2024improving}.


While there has been some work on incremental processing for increasing the fluidity of HRI \cite{hough2016investigating,abramsetal25iros}, to allow for the development of control algorithms by data-driven means, Wizard-of-Oz (WoZ) interactions, where an experimental confederate `Wizard' tele-operates a robot out of view of the user, are required. However, it is not clear to what extent current WoZ systems are capable of achieving and recording data for developing fluid HRI, nor which properties are required to achieve this.

After more precisely motivating our work and setting the contribution in context, we identify some core criteria of Wizard-of-Oz systems needed for interactive fluidity, briefly review existing WoZ-enabled robots and associated HRI experiments, specifically focusing on how they do or do not meet these criteria. We then present a prototype system in a VR environment which meets them.

\section{Motivation \& Related Work}

\begin{table*}[!t]
    \caption{An overview of existing HRI WoZ systems and their properties necessary for fluidity. ~Keys for the different interaction types: C = Controller, M = Mouse, K = Keyboard, P = Pen \& Paper, H = Haptic/Touch, G = Gesture, T = Tablet Wizarding, S = Speech X = evidenced, ? = Mentioned category but not clearly discussed, - = Not Mentioned}
    \label{tab:existing_hri_woz}
    \centering
    \resizebox{17.5cm}{!}{
    \begin{tabular}{|c|ccc|ccc|cccc|}
    \hline
        System  & 
          \rotatebox{20}{Robot name}  &                         
           \rotatebox{20}{Robot type}  &                        
            \rotatebox{20}{Real or Virtual} &                   
            \rotatebox{20}{User interaction} &                  
            \rotatebox{20}{Wizard interaction} &                
            \rotatebox{20}{Wizard interface} &                  
            \rotatebox{20}{Interruptibility and correction} &   
            \rotatebox{20}{Pollability} &                       
            \rotatebox{20}{Reproducibility} &                   
            \rotatebox{20}{Latency}                             
               \\
        
      
      \hline
  \cite{arntz2024enhancing} & Pepper & Social Robot & VR & H & M+K & Dashboard screen GUI & - & - & - & 45ms\\
  \cite{hinwood2018proposed} & UR10 & Manipulator Arm & RR & P & K & Command-line interface & X & X & - & - \\
  \cite{rietz2021woz4u} & Pepper & Social Robot & RR & ? & M+K & Dashboard screen GUI & - & - & X & \\
   \cite{runzheimer2024exploring} & CityBot & Mobile Car & VR & M+K+C & G+S & Dashboard screen GUI & - & - & X & X \\
   \cite{rodriguez2024interaction} & Citybot & Mobile Car& RR & S & M & Dashboard screen GUI & ? & - & - & - \\
   \cite{sienkiewicz2023language} & Pepper & Social Robot & RR & S + G & M & Dashboard screen GUI + Live Camera Feed & - & - & - & - \\
   \cite{arntz2024enabling} & xArm 7 & Manipulator Robot & RR & - & C & Dashboard screen GUI & X & - & - & X \\
   \cite{zou2024r2c3} & Luxai QT robot & Social robot & RR & P & T & Dashboard screen GUI & - & - & - & -\\
   \cite{zou2022Wizard} &  Luxai QT robot & Social robot & RR & P & T & Dashboard screen GUI & - & - & - & -\\
   \cite{heinisch2024physiological} & Tiago++ & Mobile Manipulator & RR & S & M+K & Dashboard screen GUI + Live Camera Feed & - & - & - & -\\
   \cite{zguda2024he} & Multiple & Social Robots & RR & - & - & - & - & - & - & -\\
   \cite{tsai2022service} & ResolutionBot & Social Robot & RR & H+S & C+M+K & Dashboard screen GUI + Live Camera Feed & - & - & - & ? \\
   \cite{schmidt2024convenience} & UR10 & Manipulator Arm & RR & S+T & M+K & - & - & - & - & -\\
   \cite{van2022Wizard} & Nao Robot & Social Robot & RR & S & ? & - & - & - & - & -\\
    \hline
    \end{tabular}
    }
    
\end{table*}

Riek~\cite{Riek2012} provides a systematic review of 54 WoZ experiments published in the primary HRI publication venues from 2001--2011. They observed that research mainly used WoZ systems for simulating the natural language processing and non-verbal action functionality of systems (72.2\% and 48.1\% of papers, respectively). They found that Wizards also performed a significant amount of navigation and mobility tasks (29.6\% of papers) while manipulation-based tasks were rarely employed (3.9\% of papers). The author hypothesises this is largely due to the difficulty of fully automating speech-based systems as well as maintaining effective tracking. Two themes that occur throughout the reviewed papers are the concern of accurately representing the natural delay that robots have within the Wizarded systems, and accounting for errors from the Wizard, intentional or otherwise. Both themes are important when considering fluidity and the potential for repairing ongoing actions. 


One shortcoming of most current WoZ systems is that they tend to employ custom Graphical User Interfaces (GUIs) fit only for their specific problem with a finite set of potential actions~\cite{martelaro2016Wizard}. The GUIs themselves can often lead to creating additional latency between a user's command and the execution of the action, as found by Runzheimer et al~\cite{runzheimer2024exploring}. 

\if
\section{Background}
To understand the state of the literature on fluidity in Wizarded HRI systems based on the challenges discussed above, we conducted a systematic review. Our research question of the literature is this: \textbf{To what extent is fluid interaction considered in the design of existing Wizard-of-Oz systems for robots?}  

\subsection{Methodology}

The reason we exclude virtual agents in general is because of our domain of interest in robotics, however we still include virtual simulations of robots. The latter exclusion criterion mainly excluded non-peer-reviewed work on personal archive sites. After these selection criteria were applied, we arrived at 14 papers where our focus was on the properties of the WoZ system they describe.

\subsubsection{Critical Appraisal}
The focus of the critical appraisal was on relevance (papers that reported an experiment, with a focus on using a Wizarded system with a robot), rigor (appropriate research describing scope, methods, execution, and research context), and credibility (conclusions based on analysis and reasoning). The critical appraisal was led by the lead researcher.

\subsection{Data Extraction}
At this stage, data from the accepted papers were extracted by reading all of the papers in detail. To gather this data, it was deposited into an Excel spreadsheet. The data collected was: The title, robot name, robot type, usage of a real or virtual robot, user modes of interaction, Wizard modes of interaction, and Wizard interface. Properties of the systems relating to fluid interaction (cf. section \ref{sec:definefluidwoz}) were added in further columns. The papers were marked in terms of the systems having these properties or not, with appropriate notes on their criteria also added.

\subsection{Synthesis}
For the evaluation of the literature, a content analysis strategy was employed, following the definition by Elo and Kyng\"{a}s~\cite{QualAnal2008}. Content analysis is a research method which provides a systematic and repeatable process~\cite{krippendorff04}. In particular, an inductive content analysis was employed which does not make any inference of meaning during the review but instead used the literal text in the paper. This approach allows to minimise potential biases in the analysis. The analysis was conducted through the lens of the criteria for fluid interaction which we defined below.
\fi

\subsection{Defining Criteria for Fluid Interaction}\label{sec:definefluidwoz}

Based on the authors' direct experiences of designing in-robot interaction control algorithms and an initial literature search, we posit the following criteria for WoZ systems to be capable of fluid interaction with users: \emph{Interruptibility and correction (IaC), Pollability, Latency measurement and optimisation}, and \emph{Reproducibility of action timings}, which are explained below.

\textbf{Interruptibility and correction}: is the ability of the Wizard to either change and update the current ongoing action of a robot to a new one, or to cancel all ongoing actions. As robots are not guaranteed to deliver perfect processing, the ability to correct the robot partway through is important for building a robust robot that people can use in real-world contexts. Additionally it is important to have a method to cancel all ongoing actions for safety reasons. In fact, safety features in personal care robots are mandatory.\footnote{https://www.iso.org/standard/53820.html} By delivering in these aspects, we can assure that we can confidently repair any incorrect actions, including those made by the wizard.


\textbf{Pollability}: refers to the definition described by Aylett and Romeo, who define something pollable as needing to know ``what state the system is in and what it has done at any point in time''~\cite{aylett2023you}. While their case focuses on speech-based robots, we widen pollability to the progress of completion of any action type. For example, if a robot like Niryo\footnote{\url{https://niryo.com/}} is used to pick and place items, the system should be pollable such that the Wizard is able to discern how far it has progressed towards completing a pick or place action at any given time, and can therefore make an optimal decision about subsequent actions, including repairing or cancelling. 

\textbf{Latency Measurement and Optimisation}: is the need for latency to be a central design concern. For effective and fluid interactions, latency must be accounted for and minimised, as even small latencies, when compounded in different parts of the system, can lead to noticeable delays. 

\textbf{Reproducibility of Action Timings}: is a WoZ system's ability to replicate how users interacted with the robot after an interaction with high timing accuracy. One of the main reasons this is important is data collection for improved future interaction -- the data gathered by accurately logging a user's actions can be either analysed, or used to train autonomous functions in the same robot and improve user experience in terms of responsiveness. 




\subsection{Review of Existing Systems}
\label{sec:results}

In an update to \cite{Riek2012}'s review, we analyse the extent to which state-of-the-art WoZ systems exhibit these 4 criteria, we conducted a literature review of recent work on WoZ systems, which resulted in 14 papers, the full results of which will be published in a follow-up paper.\footnote{The inclusion criteria which define our scope were:
\begin{itemize}
    \item The term `robot' or a reference to the type of robot was referenced in either the title or the abstract.
    \item Papers that had at least one reported study using a Wizard-of-Oz system.
    \item Papers were classified as a study and not a report, book chapter or abstract.
    \item The paper was peer-reviewed and accepted for publication.
    \item Papers were included if the Wizarded systems involved real robots or virtual simulations of the robot.
\end{itemize}
Exclusion criteria were:
\begin{itemize}
    \item Papers that focused on virtual agents without a real-world robotic equivalent instead of robots.
    \item Papers which were not available through the university services.
\end{itemize}
} Additional classifications for each system were made for \emph{Wizard and User modes of interaction}, and whether the system was \emph{Virtual or Real.} The results can be found in Table~\ref{tab:existing_hri_woz} and are described below:

\textbf{Virtual or Real:} We found the majority of studies describe WoZ systems for controlling real-world robots. \cite{runzheimer2024exploring} implemented a simulation of the CityBot and three scenes in Unity\footnote{https://unity.com/} with the user wearing the Meta Quest 2 headset.\footnote{https://www.meta.com/gb/quest/products/quest-2/} Despite the benefits of a VR simulation, the challenges of achieving fidelity to the real-world robot and environment constitute significant software design and engineering overhead, which are possible reasons for the relative lack of VR HRI systems despite their benefits. 

\textbf{Wizard and User Modes of Interaction:} There is no uniform method of designing the Wizard and User modes of interaction. For the Wizard, the use of a mouse and keyboard was the most popular mode of control, with most interfaces using a GUI dashboard with various buttons and keyboard shortcuts to execute pre-defined actions. In many cases it was not clear whether these interfaces displayed the robot's visual stream to the Wizard, or another view, such as an omniscient or third-party observation point. In many cases, it was not made clear if any action could be interrupted, cancelled, or even tracked in terms of its progression. The user interaction modes vary from haptic, to tablets to speech, depending on the affordances of the robot.


%



\textbf{Interruptibility and Correction}: If the WoZ system was reported to be able to repair current actions or cancel existing actions, it was considered to have the IaC property. Only two papers clearly achieved this. \cite{hinwood2018proposed} stated that their system had an ``interrupt node'' which allowed the Wizard to send a signal causing the robot to withdraw a safe distance to avoid a collision with the participant. It is also noted that their system had an ``override command" used to disable any running actions. The authors do not specifically mention whether the interrupt node could also be used to update the existing action.

\textbf{Pollability}:
Only one paper \cite{hinwood2018proposed} on a proposed system explicitly dealt with pollability being a core part of the system via their `robot\_feedback' publisher which was displayed back to the user. The authors specify that the 3 states they defined were `Nod', `Observe' and `Pose' and note that the Wizard could choose and control these states. However, it was unclear how the Wizard was confident in knowing how close the robot was to completing a task, or any granular information on task progress. 

\textbf{Latency:} Only three studies mention and discuss latency. \cite{runzheimer2024exploring} discusses how their Wizard interface was complicated and difficult to use, increasing the latency between a user's request and the robot commencing the action. In particular, their interface was too cumbersome for just an individual Wizard to manage alone. Their UI contained a series of buttons, sliders and input fields for scene changes, alongside having separate MIDI controllers for audio responses. Despite the Wizard being trained and having to become familiar with the system, the management of all the various controls led to delays between the commands of the participant and the wizard; especially in scenarios where the wizard did not expect an interjection from the participant leading to not so fluid interaction.


\textbf{Reproducibility of Action Timings}: Only two examples of a reproducible system for timings of action were found. \cite{runzheimer2024exploring} recorded the views and audio of both the user and the Wizard - data which was then downloadable and viewable. One particular issue they came across with their setup was that they found it impossible to effectively synchronise the various streams due to the streams coming from different sources with a variety of different settings and corrupted streams, due to their complex setup.

\begin{figure*}[!t]
\begin{center}
\caption{A diagram showing the system architecture of the proposed VR-HRI system. Hardware interfaces are blue boxes, processors are yellow, and data types are green.}
\includegraphics[width=\textwidth]{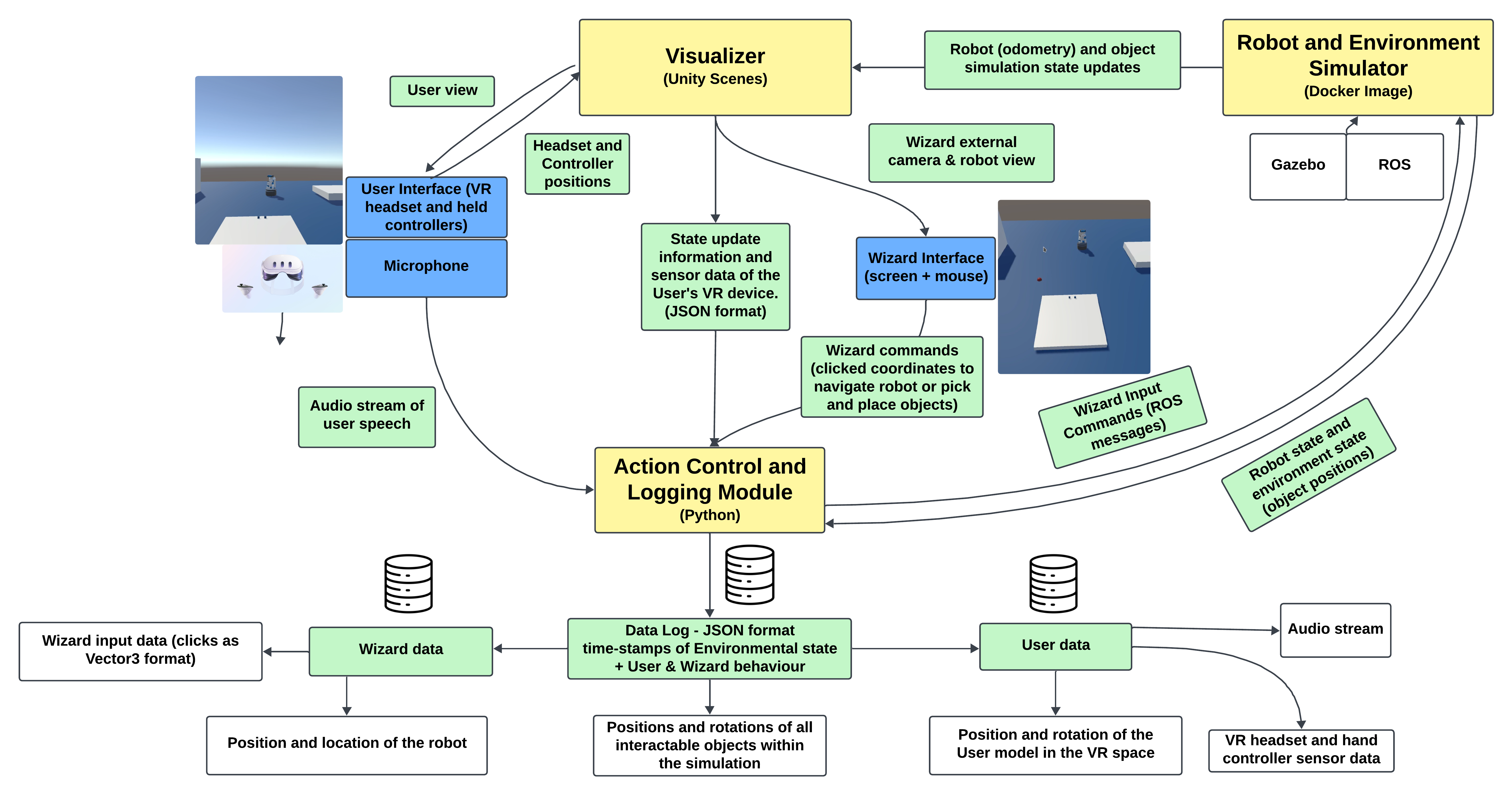}

\label{fig:architecture}
\vspace*{-0.5cm}
\end{center}
\end{figure*}

\section{Virtual Reality HRI Platform Prototype} 

As discussed above, no existing system we reviewed meets our criteria fully, with limitations ranging from high latency to over-complexity of Wizard interfaces. We have developed a test-bed WoZ system which adequately satisfies the criteria in ways which we explain below. While applicable to a variety of robots, this system uses the mobile manipulator ``Fetch'' robot\footnote{\url{https://www.zebra.com/us/en/products/autonomous-mobile-robots.html}}, controlled by speech commands by a user in the same space.\footnote{The toolkit source code is available at \url{https://github.com/fluidity-project/fluidity-hri/} and a demo video of the system's correction behaviour is at \url{https://fluidity-project.github.io/media/HRIErrors.mp4}}  

 The system architecture is shown in Figure~\ref{fig:architecture}. The ROS and Gazebo\footnote{\url{https://gazebosim.org/}}-powered \textit{Robot and Environment Simulator} is housed in a Docker container,  which sends ROS messages to the Unity-based \textit{Visualizer}. The movement and planning for our Fetch model is implemented by the ROS-based `MoveIt'\footnote{https://moveit.ai/} framework. The Visualizer, which is continuously listening for ROS messages to immediately update the relevant objects' position, provides synchronous, separate User and Wizard views of the scene. The user accesses the environment using a VR headset and hand-held controllers, while the Wizard has a separate screen to control the robot using mouse clicks. Once the Wizard has clicked on the environment for Fetch to complete a navigation or pick-and-place action, a ROS message is sent to our Python-based \textit{Action Control and Logging module}. This control module sends the request to initiate the action back to the Robot and Environment Simulator for execution. 
 

To be able to receive ROS messages in Unity, we use a ROS-TCP connector\footnote{\url{https://github.com/Unity-Technologies/ROS-TCP-Connector}} and to receive ROS messages back from Unity we use a ROS-TCP-Connector package in the Docker container\footnote{\url{https://github.com/Unity-Technologies/ROS-TCP-Endpoint}}. To keep the robot visualisation consistent, we use the Fetch UDRF descriptor file and meshes and import the robot model into the Unity environment using the UDRF-Importer package\footnote{\url{https://github.com/Unity-Technologies/URDF-Importer}}.
 
\textbf{Wizard Interface:} The Wizard sees the simulated robot camera feed on the Fetch model from the Visualizer and has full continuous audio access to the User's microphone. Responding as fast as possible to user commands, the Wizard mouse-clicks on a target destination for moving and selecting objects and destinations. Once this is registered, a ROS message is sent to our control module. Visualisations signal the Wizard's current target location (red cone), target object (green highlight) and/or object destination (blue spot on surface). There is a simple dashboard with a `Cancel all actions' button and a message area to signal illegal clicks.

\textbf{User Interaction:} The user wears a VR headset displaying the Unity Scenes and a microphone to interact with the (Wizarded) robot via speech. They also have limited movement capabilities through the use of hand-held controllers to navigate the environment.\footnote{All Unity-compatible hardware can be used. We use the Meta Quest 3 VR headset and Meta Quest Touch Plus -~\url{https://www.meta.com/gb/quest/quest-3/}} All sensor data is sent to the control and logging module. 

\textbf{Logging data:} In addition to the audio feed from the user's microphone being recorded, all the following data is logged and timestamped in a JSON format (with timestamps and an ID tag): 1 ) Wizard actions: time-stamped vectors of mouse clicks. 2) The robot's state: position of its base and joint state. 3) Interactable objects within the space: positions and rotations. 4) The user's view: position of the camera in the Unity scene as controlled by the VR headset. 5) User's controller actions. In playback mode, log files can be read in by the system to simulate the interaction.

\subsection{Discussion: Meeting of Fluidity Criteria}
While the immediate next step in our future work is evaluating the system with real users, here we discuss how our VR WoZ prototypes meets our criteria identified in \ref{sec:definefluidwoz}.

\textbf{Interruptibility and Correction} is met in our system by 1) Goal update functionality, which allows the Wizard to update the current action of the robot by a simple override with a button click on a new location or object; 2) The \textit{cancel} button, which allows us to revoke all current goals to stop the robot from completing any remaining actions and come to a default state as quickly as possible, modulo deceleration.

\textbf{Pollability} is met as the Wizard can directly see the progress of the action through their view, and the data log records the state of the robot during any given action. 

\textbf{Latency Measurement and Optimization} is achieved by using ROS nodes which tether the robot simulation to both User and Wizard views. The system can also attribute its overall latency to its various causes: 1) The delay from the user requesting an action via speech to the Wizard initiating that action. 2) The delay from the Wizard's perception of the action request to the action starting. 3) The delay from the action starting to the user perceiving the start of that action. 4) The delay from interrupting or repairing any given action. For minimising 1) and 3) we allow full access to the audio stream in the Wizard's interface and continuous visual access to the synchronous Unity views. We minimise 2) by experimenting with different automatic recovery behaviours from the Fetch MoveIt package, which are shown to have a large effect on user perceptions of the robot's abilities \cite{schulz2021movement}, here making them less conservative to lower the latency of task execution. For 4), fluid repair is achieved by the Wizard being able to select new target destinations and objects during ongoing motion, similar to \cite{abramsetal25iros}.

\textbf{Reproducibility} is achieved by the accurate time-stamped logging of the data described above. While this data does not exhaustively include all visual information, full reconstruction of the User and Wizard behaviour and the changing states of the environment is possible, as are real-time User, Wizard and third-party views.

\section{Conclusion and Future Work}
Based on requirements for fluid HRI for WoZ-controlled robots, we have presented a prototype VR platform which allows for the functional properties of Interruptibility and correction, pollability, latency measurement and optimisation, and reproducibility of action timings. Our toolkit is generalizable beyond the Fetch robot to any ROS-based simulation and Unity environment, allowing customizable worlds and robots. In future work, we will run a user study exploring how effective the system is in facilitating fluid interactions both in WoZ mode and in autonomous mode experiments with systems trained on the WoZ-collected data.

\section*{Acknowledgements}
Baptista de Lima, Hough and F{\"o}rster's work is funded by the FLUIDITY project (EPSRC grant EP/X009343/1).

\bibliographystyle{ACM-Reference-Format}
\bibliography{refs}

\appendix

\end{document}